\documentclass[sigconf]{acmart}
\usepackage{graphicx}
\usepackage{amsmath}
\usepackage{booktabs}
\usepackage{makecell}
\usepackage{multirow}
\usepackage{pifont}
\usepackage{wrapfig}
\usepackage{graphicx}
\usepackage{xcolor}    
\usepackage{bm}
\AtBeginDocument{%
  }

\setcopyright{acmcopyright}
\copyrightyear{2023}
\acmYear{2023}
\setcopyright{acmlicensed}\acmConference[MM '23]{Proceedings of the 31st ACM International Conference on Multimedia}{October 29-November 3, 2023}{Ottawa, ON, Canada}
\acmBooktitle{Proceedings of the 31st ACM International Conference on Multimedia (MM '23), October 29-November 3, 2023, Ottawa, ON, Canada}
\acmPrice{15.00}
\acmDOI{10.1145/3581783.3611745}
\acmISBN{979-8-4007-0108-5/23/10}

\acmSubmissionID{235}

\begin{document}

\title{AdaBrowse: Adaptive Video Browser for Efficient Continuous Sign Language Recognition}


\author{Lianyu Hu}
\affiliation{%
  \institution{Tianjin University, China}
  \country{}}
\email{hly2021@tju.edu.cn}

\author{Liqing Gao}
\affiliation{%
\institution{Tianjin University, China}
\country{}}
\email{lqgao@tju.edu.cn}

\author{Zekang Liu}
\affiliation{%
\institution{Tianjin University, China}
\country{}}
\email{lzk100953@tju.edu.cn}

\author{Chi-Man Pun}
\affiliation{%
  \institution{University of Macau, China}
  \country{}}
\email{cmpun@umac.mo}

\author{Wei Feng}
\affiliation{%
\institution{Tianjin University, China}
\country{}}
\email{wfeng@ieee.org}
\renewcommand{\shortauthors}{Lianyu Hu, Liqing Gao, Zekang Liu, Chi-Man Pun, Wei Feng}

\begin{abstract}
Raw videos have been proven to own considerable feature redundancy where in many cases only a portion of frames can already meet the requirements for accurate recognition. In this paper, we are interested in whether such redundancy can be effectively leveraged to facilitate efficient inference in continuous sign language recognition (CSLR). We propose a novel adaptive model (AdaBrowse) to dynamically select a most informative subsequence from input video sequences by modelling this problem as a sequential decision task. In specific, we first utilize a lightweight network to quickly scan input videos to extract coarse features. Then these features are fed into a policy network to intelligently select a subsequence to process. The corresponding subsequence is finally inferred by a normal CSLR model for sentence prediction. As only a portion of frames are processed in this procedure, the total computations can be considerably saved. Besides temporal redundancy, we are also interested in whether the inherent spatial redundancy can be seamlessly integrated together to achieve further efficiency, i.e., dynamically selecting a lowest input resolution for each sample, whose model is referred to as AdaBrowse+. Extensive experimental results on four large-scale CSLR datasets, i.e., PHOENIX14, PHOENIX14-T, CSL-Daily and CSL, demonstrate the effectiveness of AdaBrowse and AdaBrowse+ by achieving comparable accuracy with state-of-the-art methods with 1.44$\times$ throughput and 2.12$\times$ fewer FLOPs. Comparisons with other commonly-used 2D CNNs and adaptive efficient methods verify the effectiveness of AdaBrowse. Code is available at \url{https://github.com/hulianyuyy/AdaBrowse}.
\end{abstract}




\maketitle

\section{Introduction}
Sign language is one of the most important communication tools for the deaf people in their daily life. However, mastering this language is rather difficult and exhaustive for the hearing people, which severely hinders direct communication between the two groups. Continuous sign language recognition (CSLR) aims to automatically translate input sign videos into sentences, which may greatly relieve this dilemma and provide more non-intrusive communication channels for both groups. Despite great progress achieved by recent methods~\cite{Min_2021_ICCV,pu2019iterative,niu2020stochastic,zhou2020spatial,cui2019deep,cheng2020fully,pu2020boosting,koller2017re,niu2020stochastic} upon accuracy in CSLR, they typically bring plenty of computations and cause a low throughput, which is especially undesirable towards real-life scenarios with demands of low computations and real-time processing. 
  
In this paper, we aim to address this issue by leveraging the inherent feature redundancy in raw videos to save unnecessary computations, based on the observation that not all frames are equally important for CSLR. On the one hand, the shot video is usually information-intensive. One gloss\footnote{Gloss is the atomic lexical unit to annotate sign languages.} in a sign video always consumes tens of frames to record. However, our human beings could scan just few frames to recognize a sign, which shows that the involved information in videos is inherently redundant. Serially processing all frames may waste a lot of computations. On the other hand, we notice that different videos may require information of different degrees for recognition. For example, some sign videos are expressed with large body movements with few disturbances, which could be recognized with low resolutions easily. In contrast, the signs of some videos are quite similar which require detailed spatial features of high resolutions to recognize. The required computations are thus unevenly distributed across different videos. Thus, equally treating all videos with full images (224$\times$224) may be unnecessary.
  
\begin{figure}[t]
    \centering
    \includegraphics[width=\linewidth]{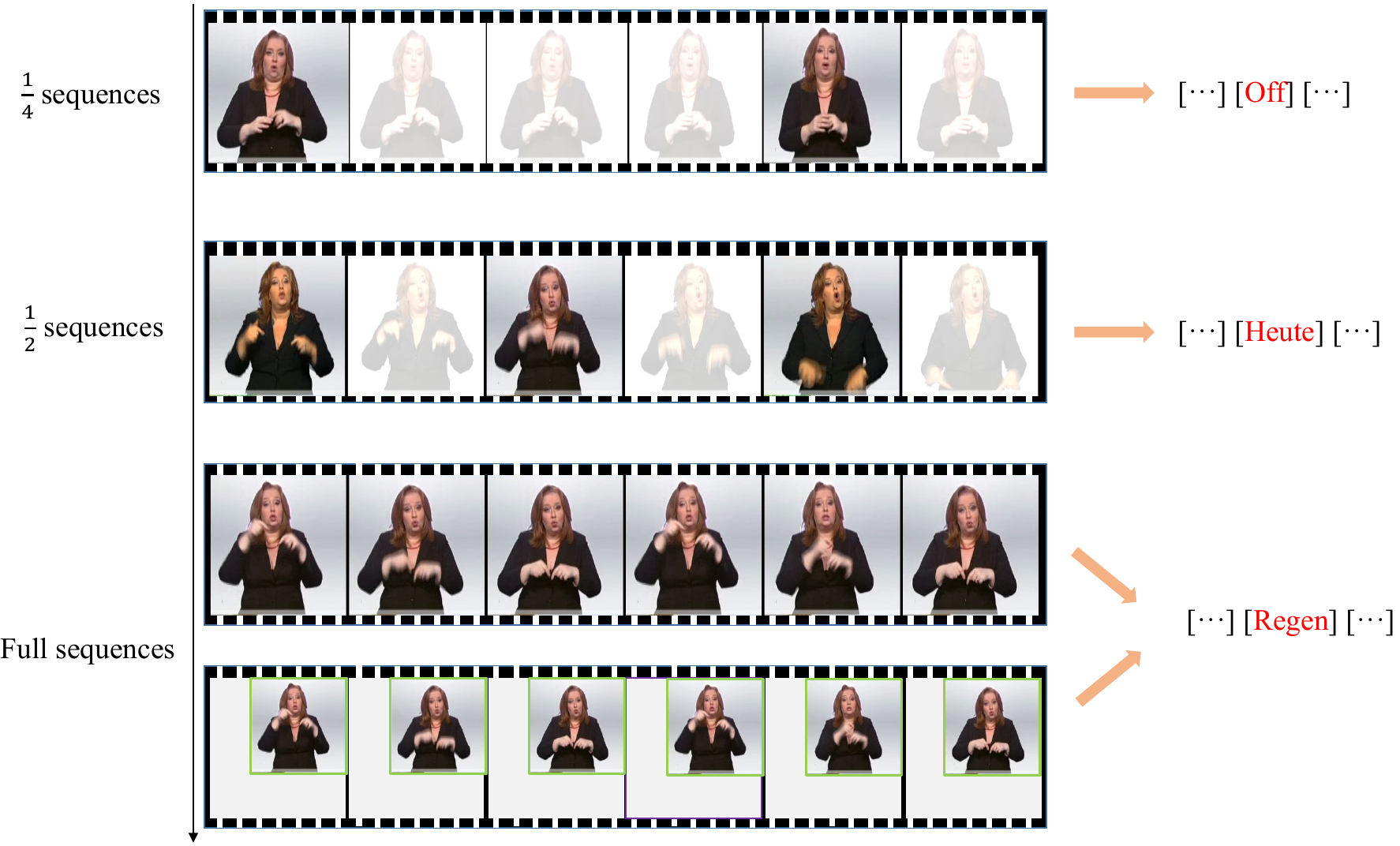} 
    \caption{Videos in many cases can be accurately recognized from only a portion of frames or from small inputs (bottom).}
    \label{fig1}
  \end{figure}

In this paper, we present AdaBrowse to leverage spatial and temporal feature redundancy to achieve efficient CSLR. For the temporal domain, we propose to dynamically select a subsequence of input videos to process to save unnecessary computations. As shown in fig.~\ref{fig1}, in our experiments we found "easy" videos of relatively small human movements and sparse spatial details, usually require fewer frames (e.g., $\frac{1}{4}$ and $\frac{1}{2}$). In contrast, "hard" videos always require an intact input sequence to fully explore frame correlations to reduce mistakes as much as possible. In specific, our model first utilizes a lightweight global convolutional neural network (CNN) to take a quick glance at input videos to extract cheap and coarse features. Then a recurrent policy network is trained based on these preceding features to select a most informative subsequence under limited computational costs. This procedure refers to the Gumbel-Softmax Algorithm~\cite{jang2016categorical} to tackle the non-differentiable problem resulting from discrete sampling from multiple subsequence candidates. Finally, we feed the corresponding subsequence into an attached CSLR model to perform sentence prediction. As only a portion of frames are processed in this procedure, the total computations can be considerably saved. For the spatial domain, we propose to adaptively switch the input resolution for each video to reduce unnecessary computations. This is based on the observation that many videos can be perfectly recognized in sequences of relatively small images (the bottom of fig.~\ref{fig1}). This is achieved by naturally extending the policy network in AdaBrowse into a unified form to both determine input resolution and select a subsequence for processing, whose model is referred to as AdaBrowse+. As we show in the experiments, AdaBrowse series can intelligently learn to unevenly allocate computing resources among multiple subsequence candidates to balance between accuracy and computations.
  
Experimental results on four widely-used large-scale datasets, i.e., PHOENIX14, PHOENIX14-T, CSL-Daily and CSL, demonstrate AdaBrowse by itself can reduce a number of FLOPs\footnote{FLOPs denotes the number of multiply-add operations and GFLOPs denotes measuring FLOPs by giga.} and achieve better throughput. AdaBrowse+ further promotes the efficiency to 2.12$\times$ fewer FLOPs and 1.44$\times$ better throughput. Comparisons with other commonly-used 2D CNNs and adaptive efficient methods verify the effectiveness of AdaBrowse. Notably, AdaBrowse series could achieve comparable accuracy with other heavy state-of-the-art methods, with much lower computations. Various in-depth analyses and visualizations are given to comprehensively show their effects from various perspectives.
  
\section{Related Work}
\subsection{Continuous Sign Language Recognition}
Sign Language Recognition methods can be roughly categorized into isolated Sign Language Recognition (ISLR) and Continuous Sign Language Recognition (CSLR), and we focus on the latter one in this paper. Different from ISLR classifying a video into a single label, CSLR is a sequence-to-sequence task by translating input videos into target sentences, which is a potentially valuable tool for automatic sign video translation in real-life scenarios. Earlier methods~\cite{gao2004chinese,han2009modelling,freeman1995orientation,koller2015continuous} in CSLR usually rely on hand-crafted features or Hidden Markov Model-based systems~\cite{koller2016deepsign,koller2017re,koller2016deep} to split videos and then perform translation. The recent success of CNNs and RNNs brings great progress for CSLR. Equipped with CTC loss~\cite{graves2006connectionist} to align target sentences with predicted sentences for supervision, they~\cite{cihan2017subunets,cui2017recurrent,cui2017recurrent,pu2019iterative,cui2019deep,cheng2020fully,Min_2021_ICCV} usually first employ a 2D CNN to extract frame-wise features, and then deploy hybrids of 1D CNNs and LSTM to capture temporal dependencies, followed by a classifier to perform sentence prediction. However, some methods~\cite{cui2017recurrent,pu2019iterative,cui2019deep} find that the 2D CNN is not well-trained in such a paradigm and propose to reuse the iterative training strategy to refine it, causing much longer training time and bringing more computations. More recent studies attempt to relieve this problem by directly enhancing the 2D CNN with extra modules and visual losses and~\cite{cheng2020fully, Min_2021_ICCV,hao2021self} or squeezing beneficial temporal information~\cite{hu2022temporal}. Our method is orthogonal to and can be flexibly deployed upon these methods for efficient inference with preserved accuracy.
  
\subsection{Reducing Temporal Redundancy}
The inherent temporal redundancy in videos has been widely explored in related tasks based on the observation that not all frames contribute equally for recognition. Thus, a number of computations can be saved by paying less/no attention to less informative frames. These related methods can be roughly divided as follows: (1) early stopping~\cite{ghodrati2021frameexit,fan2018watching}, where these methods propose to adaptively terminate the computation once a target prediction confidence threshold is met to save the following computations; (2) frame sampling~\cite{wang2021adaptive,wu2019multi,gao2020listen,korbar2019scsampler,wu2020dynamic}, i.e., dynamically selecting a portion of input frames to process and discarding the rest; (3) conditional computing~\cite{wu2019liteeval,meng2020ar}, e.g., conditionally extracting coarse or expensive features at each timestep through a decision-maker to save unnecessary computations over less important frames. Besides the above-mentioned methods, Adafuse~\cite{meng2021adafuse} proposes to dynamically decide whether to reuse features from past timesteps to save computations. Our AdaBrowse differentiates from these methods in a novel way to reduce temporal redundancy by selecting one optimal subsequence to process, with an implicit assumption that there is always a shortest subsequence containing most necessary information for recognition in each video.
  
\subsection{Reducing Spatial Redundancy}
Spatial redundancy has been broadly visited by previous methods in 2D image domain to reduce computations. Based on the observation that easy samples can be recognized over relatively smaller inputs while hard samples typically require larger inputs to explore finer details, considerable computations can be reduced by unevenly distributing computations among samples. Such related implementations include (1) resolution switcher~\cite{yang2020resolution,meng2020ar,DBLP:conf/nips/ZhuHWZNLW21}, i.e., dynamically switching between different input resolutions to provide confident enough predictions for each sample under limited computational costs; (2) attending to task-relevant regions~\cite{xie2020spatially,wang2020glance,figurnov2017spatially}, e.g., only focusing on the small task-relevant regions for recognition while stopping processing other irrelevant spatial regions for computation savings. Despite great progress in 2D image domain, such spatial redundancy has been relatively few visited in video tasks~\cite{wang2021adaptive,meng2020ar}, especially in CSLR, while we are interested in whether it can be further combined with temporal redundancy for higher efficiency. 
  
\section{Method}
Inspired by the fact that videos can be perfectly translated into correct sentences with only a small portion of frames in a lot of scenarios, we seek to save the computations spent on redundant frames for efficiency. Furthermore, we notice the expensive computations spent on fixed high-resolution input (e.g., 224$\times$224) may not always be necessary. For example, easy samples can be easily distinguished with small blurry frames (e.g., 96$\times$96) while hard samples may require fine details contained in high-resolution inputs. To achieve efficient inference, we first propose an adaptive framework, coined as AdaBrowse, to only process partial inputs by leveraging the temporal redundancy, where considerable computational costs can be saved without sacrificing accuracy. We further update it by introducing the inherent spatial redundancy in images by considering frame resolution, referred to as AdaBrowse+.   
  
  
  \begin{figure*}[t]
    \centering
    \includegraphics[width=\textwidth]{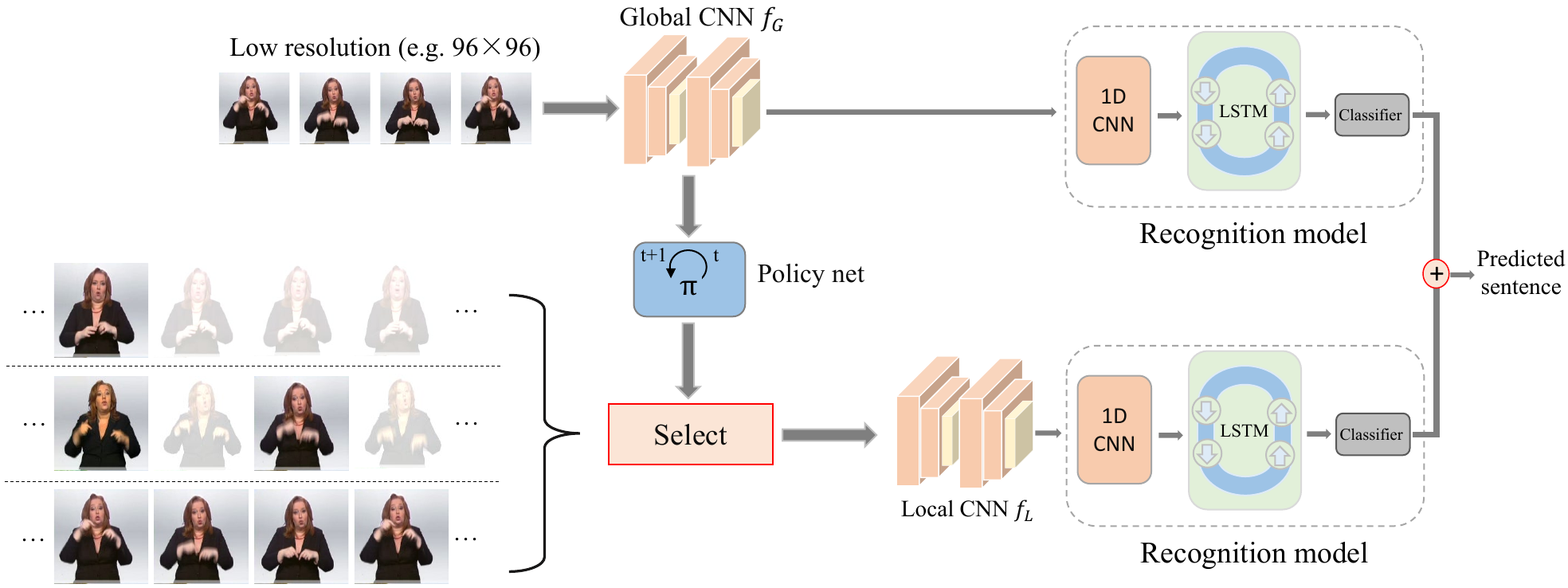} 
    \caption{An overview for AdaBrowse. A lightweight Global CNN $f_G$ is first employed to extract coarse features. These features are then sent into a policy network $\pi$ to select a target subsequence which is finally inferred by an attached recognition model. Features extracted by $f_G$ are reused and inferred by another lightweight recognition model and outputs from two branches are averaged as final predictions.}
    \label{fig2}
  \end{figure*}
    
\subsection{AdaBrowse}
\subsubsection{Overview.} We first give an overview of AdaBrowse in fig.~\ref{fig2}. Given an input sequence, AdaBrowse needs to dynamically select a suitable input sequence length for the attached CSLR model to maximize accuracy under limited computational budgets. For this goal, a Global CNN $f_G$ is first utilized to take a quick glance at the input sequence, obtaining coarse and cheap features. For efficient inference, this procedure is required to be as lightweight as possible. Thus, we let the Global CNN $f_G$ scan input of relatively lower resolutions (e.g., 96$\times$96) which only consumes small extra computations. Then, these features are fed into a policy network $\pi$ to holistically consider information across frames to select an appropriate input length from $M$ subsequence candidates. Once the decision is made, a corresponding subsequence will be selected and fed into a common Local 2D CNN $f_L$ to extract frame-wise features followed by a recognition model to predict sentences. The features extracted by $f_G$ will be reused and inferred by a lightweight recognition model, whose outputs are averaged with those from another branch as final predictions. We will detail these components in the following.
\subsubsection{Global CNN $f_G$ and Local CNN $f_L$}
The Global CNN $f_G$ is utilized to rapidly process input frames for a quick glance to obtain coarse and cheap global features. Compared to the common Local 2D CNN $f_L$ used to extract expensive but powerful features for the following recognition model, $f_G$ has to be circularly executed for each sample and thus is required to consume as few computations as possible. Therefore, we fed frames of relatively small resolutions (e.g., 96$\times$96) into $f_G$ to make it quite lightweight and fast, which can still provide reliable global features for the subsequent policy network $\pi$ to make decisions. Instead, we deploy $f_L$ as a powerful backbone with high-resolution inputs to offer discriminative representations for the recognition model. 
  
Specifically, given $T$ input frames $V$=\{ $v_1$, $\dots$, $v_T$\}, $f_G$ will take the resized low-resolution frames as inputs and produce coarse features $x_t^G$ as :
  \begin{equation}
    \label{e1}
      x_t^G = f_G({\rm Resize}(v_t)),\quad t\in\mathbb{R}^{T}.
    \end{equation} 
Instead, $f_L$ will process a selected target subsequence into discriminative representations as:
  \begin{equation}
    \label{e2}
      x_{t}^L = f_L(\widetilde{v}_{t}),\quad t\in\mathbb{R}^{T}
    \end{equation} 
where $\widetilde{v}$ is the selected subsequence by the policy network $\pi$ introduced next.
  
\subsubsection{Policy network $\pi$}
The policy network $\pi$ receives global coarse features $x^G$ from $f_G$ to select a suitable length for each input sequence to process. The policy network should receive both past and future information to help infer the suitable input subsequence for each video. Besides, it's better to browse the input video backward and forward to obtain an overall perspective of the input video. Thus, we instantiate the policy network $\pi$ as a GRU~\cite{cho2014learning} to aggregate cross-frame information, which is followed by a MLP to predict which subsequence to use. The outputs of the MLP is a vector $c$ with size of $\mathbb{R}^{2^M-1}$, which predicts the distribution of selecting $2^M$-1 subsequences.

For the subsequence generation, we divide the input into subsequences with different lengths beforehand. Formally, with predefined $M$ kinds of subsequence length, we sample inputs with interval \{1,$\dots$,$2^{M-1}$\}. The resulted subsequences own \{1,$\dots$,$\frac{1}{2^{M-1}}$\} length compared to the inputs. To enrich the action representation diversity, for $m_{th}$ subsequence, we sample it with different starting and ending offsets by \{$1,\dots,2^{m-1}$\}, resulting in $2^{m-1}$ subspecies. Thus, there would be totally \{$1+2^1+\dots,2^{M-1}$\}= $2^{M}-1$ subsequence candidates. 

Specifically, as $\pi$ selects a subsequence from $2^M-1$ candidates, it leads to a non-differentiable problem. We refer to the Gumbel-Softmax Algorithm~\cite{jang2016categorical} to solve this discrete sampling issue. In specific, the choice $c\in \mathbb{R}^{2^M-1}$ of the selected subsequence is drawn from the distribution:
  \begin{equation}
    \label{e3}
      c \sim \pi(\cdot|x^G, h^{\pi})
    \end{equation} 
where $h^{\pi}$ is the hidden state of the final timestep in GRU.
Once the choice $c$ is generated, it will be transposed into a one-hot vector and multiplied with subsequence candidates to select a target one for further processing.

\subsubsection{Gumbel-Softmax Algorithm}
We refer to the Gumbel-Softmax Algorithm\cite{jang2016categorical} to help resolve the non-differentiable problem in sampling $c$ from a discrete distribution, and make it differentiable to allow backward gradient flow. In specific, given the normalized logits $z \in \mathbb{R}^{M}$ generated by $\pi$ for predicting $c$, the Gumbel-Softmax Algorithm adds Gumbel noise $g \in \mathbb{R}^{M}$ to $z$ and then draw a discrete sample:
  \begin{equation}
  \label{e4}
    \tilde{z}=\mathop{argmax}\limits_{i \in M}(z^i+g^i).
  \end{equation}
Here $g^i=-log(-logU_i)$ represents a standard Gumbel distribution with $U_i$ sampled from a uniform i.i.d distribution $U(0,1)$. $\tilde{z}$ is used to choose frame resolution in the network forward pass. Due to the non-differentiable property of argmax operation in Eq.~\ref{e4}, the Gumbel-Softmax distribution is used as a continuous relaxation for argmax to allow back propagating from the discrete sample. In specific, the one-hot coding of $\tilde{z}$ is relaxed to a normalized distribution $\hat{z}$ using softmax:
  \begin{equation}
  \label{e5}
    \hat{z^i}=\frac{exp((log(z^i)+g^i)/\tau)}{\sum_{i=0}^C exp((log(z^i)+g^i)/\tau)}
  \end{equation}
where $\tau$ is a temperature parameter that controls the approximation degree of $\hat{z}$ to $\tilde{z}$. When $\tau \to +\infty$, $\hat{z}$ becomes uniform distribution. While $\tau \to 0$, $\hat{z}$ approaches a one-hot vector i.e., $\tilde{z}$. In practice, $\tau$ is always set as a big value and then annealed down to approaching 0 as training progresses.

\subsubsection{Recognition model}
The recognition model receives preceding representations $x_L$ from $f_L$ and $x_G$ from $f_G$ to produce predicted sentences. We set the recognition model as a series of 1D CNN, a BiLSTM and a fully connected layer, following state-of-the-art CSLR methods~\cite{Min_2021_ICCV,hao2021self}. In this sense, our model can be viewed for fair comparison with current CSLR methods, which only equips with additional lightweight $f_G$, $\pi$ and another recognition model to adaptively make decisions.
  
\subsection{Training Algorithm}
\subsubsection{Training procedure}
To better cooperate different components in AdaBrowse with each other, we present a two-stage training algorithm.
  
\textbf{Stage \uppercase\expandafter{\romannumeral1}: Warm-up.} We first equip $f_G$ and $f_L$ with the recognition model $e_G$ and $e_L$ over their corresponding input resolutions to prepare well-initialized parameters separately, by training with the standard CTC loss~\cite{graves2006connectionist} over the training set $D_{train}$: 
  \begin{equation}
    \label{e6}
    \mathop{minimize}\limits_{f_L, e_L, f_G, e_G} \quad \mathbb{E}_{V_i\in D_{train}}\left[\frac{1}{N}\sum_{i=0}^N \mathcal{L}_{CTC}(p_i, y_i).\right] 
    \end{equation}
where $p_i$ and $y_i$ denotes the prediction and ground truth label of $V_i$ in $D_{train}$, respectively, and $i$ is the index number.   
  
\textbf{Stage \uppercase\expandafter{\romannumeral2}: Cooperation.}
Given the well-initialized parameters for $f_G$, $f_L$, $e_G$ and $e_L$, we cooperate them within a whole framework. Specifically, a low-resolution input is first fed into $f_G$ to produce coarse features, which will both undergo $\pi$ to provide decisions $c$ and get through $e_G$ to obtain sentence predictions $p^G$. Then features of $2^M-1$ subsequence candidates are multiplied with the transposed one-hot vector of $c$ to produce a selected subsequence, which will be sent into $f_L$ and $e_L$ to produce sentence predictions $p^L$. Finally, $p^G$ and $p^L$ will be averaged to give final outputs. The training goal can be presented as: 
  \begin{equation}
    \begin{aligned}
    \label{e7}
    \mathop{minimize}\limits_{f_L, e_L, f_G, e_G, \pi} \quad \mathbb{E}_{V_i\in D_{train}}   \frac{1}{N}\sum_{i=0}^N \left[ \mathcal{L}_{CTC}(p^G_i, y_i) \notag\right. \\  
    \left.  + \mathcal{L}_{CTC}(p^L_i, y_i) + \alpha \mathcal{L}_{{\rm Effi}} + \beta \mathcal{L}_{{\rm Align}}. \right]
    \end{aligned}
    \end{equation}
  Here, $\mathcal{L}_{{\rm Effi}}$ and $\mathcal{L}_{{\rm Align}}$ denote the efficiency loss and alignment loss, with $\alpha$ and $\beta$ representing their weights, respectively. We will introduce them next.
\subsubsection{Loss functions}
Except for the standard CTC Loss~\cite{graves2006connectionist} to align target sentences with predicted sentences, we propose another two loss functions $\mathcal{L}_{{\rm Effi}}$ and $\mathcal{L}_{{\rm Align}}$ to balance between accuracy and computational costs, and coordinate them with the network components in AdaBrowse for additional supervision.

\textbf{CTC Loss} $\mathcal{L}_{{\rm CTC}}$. It aligns an input video $\mathcal{V}$ = ($v_1$, $\cdots$, $v_T$) with a target sentences with $U$ glosses, $\mathcal{G}$ = ($g_1$, $\cdots$, $g_U$). Specifically, CTC loss introduces a blank label (-) to represent unlabeled data (non-gesture segments or transitional frames) in the target sentence. Then CTC loss builds a many-to-one function $\mathcal{B}$, to align input frames with output glosses referred to as path $l$ by removing the repeated and blank label in the target sentence, e.g. $\mathcal{B}$(-S-u-n-n-d-a-a-y-) = $\mathcal{B}$(-S-u-n-d-a-y-) = Sunday. With the help of this mapping function $\mathcal{B}$, CTC loss provides end-to-end supervision for the parameters of the model by summing the probabilities of all feasible paths:
\begin{equation}
  \label{e8}
  \begin{aligned}
    \mathcal{L}_{{\rm CTC}} &= -{\rm log}\; p(\bm{\mathcal{G}}|\bm{\mathcal{V}};\theta) \\ &= -{\rm log}(\sum_{l\in \mathcal{B}^{-1}(\bm{\mathcal{G}})} p(l|\bm{\mathcal{V}};\theta) ).
  \end{aligned}
\end{equation}
Based on the conditional independence assumption, the conditional probability $p(l|\bm{\mathcal{V}})$ can be calculated as follows: 
\begin{equation}
  \label{e9}
  p(l|\bm{\mathcal{V}};\theta) ) = \prod_{t=1}^{T}{p(l_t|\mathcal{V};\theta)}
\end{equation}
where the probabilities are exactly the normalized output logits generated by the network classifier.

\textbf{Efficiency loss $\mathcal{L}_{{\rm Effi}}$.} Specifically, we precalculate a computation table $s\in \mathbb{R}^{2^M-1}$ in advance to save the computations of processing $2^M-1$ subsequences. The efficiency loss $\mathcal{L}_{{\rm Effi}}$ can be computed by multiplying the one-hot vector \textit{c} with \textit{s}, to query the computations of as:
  \begin{equation}
  \label{e10}
  \mathcal{L}_{{\rm Effi}} =  c \times s.
  \end{equation}
It can be set as different values to produce a set of models under various demands.
  
\textbf{Alignment Loss $\mathcal{L}_{{\rm Align}}$.} Considering $f_L$ and $f_G$ are pretrained over different resolutions, the representations $x_G$ generated by $f_G$ will mostly be less powerful than $x_L$ generated by $f_L$. We propose $\mathcal{L}_{{\rm Align}}$ to further enhance $f_G$ to provide more stable decisions $c$ and better sentence predictions $p^G$ by distilling valuable features from $x_L$ to $x_G$ with Kullback-Leibler divergence as:
  \begin{equation}
    \label{e11}
    \mathcal{L}_{{\rm Align}} = {\rm KL}({\rm softmax}(\frac{x^G}{\gamma }),{\rm softmax}({\frac{x^L}{\gamma}}))
    \end{equation}
where $\gamma$ controls the "soften" degree and is set as 8.
  
\subsection{Reducing Spatial Redundancy}
Except for temporal redundancy, we notice that lots of "easy" videos can be perfectly recognized over smaller inputs (low resolution), where identically feeding the model with high-resolution frames may be unnecessary, and inevitably wastes plenty of computations. We propose an extended version of AdaBrowse, coined as AdaBrowse+, by incorporating frame resolution decision into our framework.
  
  \begin{figure}[t]
    \centering
    \includegraphics[width=\linewidth]{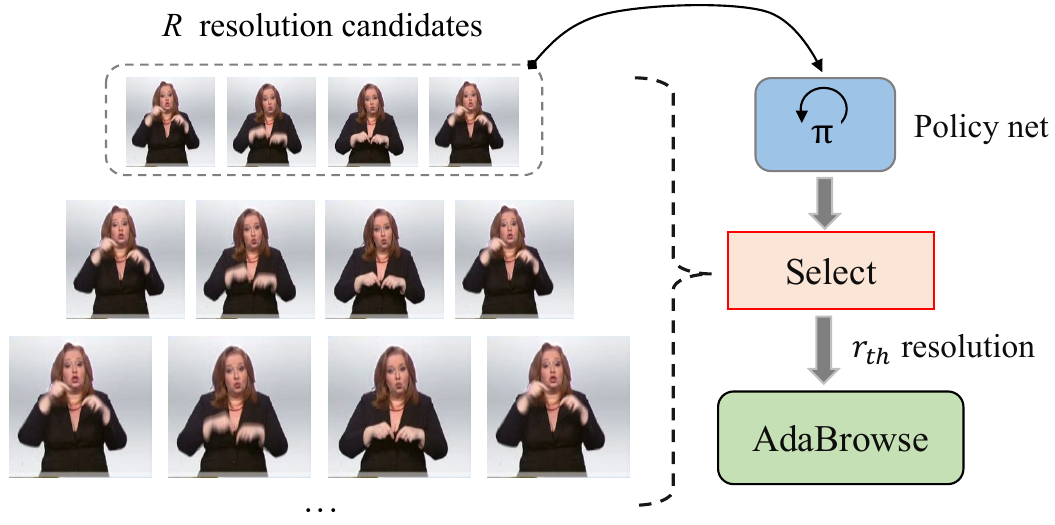} 
    \caption{An overview for the updated part in AdaBrowse+.}
    \label{fig3}
  \end{figure}
 
Given partitioned subsequences, we further divide each subsequence into $R$ different resolutions from the lowest to the highest (normal input), to be holistically aware of spatial-temporal redundancy. With $R$ resolution candidates, we always feed the lowest resolution into $f_G$ to extract coarse features $x_G$ with the goal of saving computations. $x_G$ will be sent into $\pi$ to make decisions over both resolutions and subsequences. As a full sequence of the lowest resolution has been processed, it won't be further divided into shorter subsequences and thus the number of all candidates is $(R-1)\times (2^M-1)$+1. The precalculated computation table $s$ in $\mathcal{L}_{{\rm Effi}}$ is expanded as well.
  
\section{Experiments}
\subsection{Experimental Setup}
\textbf{Datasets.}
\textbf{PHOENIX14}~\cite{koller2015continuous} and \textbf{PHOENIX14-T}~\cite{camgoz2018neural} are both recorded from the German TV weather forecasts. All actions are performed by nine actors wearing dark clothes in front of a clean background. They contain 6841/8247 different sentences with a vocabulary of size 1296/1085 which are split into 5672/7096 training samples, 540/519 development (Dev) samples and 629/642 testing (Test) samples. \textbf{CSL-Daily}~\cite{zhou2021improving} revolves the daily life, recorded indoors at 30fps by 10 signers. It contains 20654 sentences, divided into 18401 training samples, 1077 development (Dev) samples and 1176 testing (Test) samples. \textbf{CSL}~\cite{huang2018video} is recorded in the laboratory by 15 signers with a vocabulary size of 178 with 100 sentences. 25000 videos are divided into training/testing sets by 8:2. 
  
  \begin{table*}[t]
    \setlength\tabcolsep{5pt}
    \centering
    \caption{Results for AdaBrowse and AdaBrowse+ on the PHOENIX14, PHOENIX14-T, CSL-Daily and CSL datasets. Throughput is measured on a 3090 graphical card with data cached and batch size 1.}
    \begin{tabular}{ccccccccccc}
      \hline
      \multirow{2}{*}{Methods} & \multirow{2}{*}{$\alpha$} & \multirow{2}{*}{\makecell[c]{GFLOPs}} & \multirow{2}{*}{\makecell[c]{Throughput \\ (videos/s)}} & \multicolumn{2}{c}{\textbf{PHOENIX14}} & \multicolumn{2}{c}{\textbf{PHOENIX14-T}} & \multicolumn{2}{c}{\textbf{CSL-Daily}} & \multirow{2}{*}{\textbf{CSL}}\\
      &  &  & & Dev(\%)    & Test(\%)     & Dev(\%)    & Test(\%)   & Dev(\%)    & Test(\%) &  \\
      \hline
      Baseline & - & 364  & 12.22 & 19.7  &20.9  & 19.5 & 20.8 & 31.5 & 30.8 & 0.9  \\               
      AdaBrowse & 0.04  &  287    & 12.31  & 19.4 & 20.5 & 19.3 & 20.5  & 31.1 & 30.6 & 0.6\\
      AdaBrowse  & 0.15 &  254    & 14.26  & 19.6 & 20.8 & 19.4 & 20.6  & 31.3 & 30.6 & 0.8      \\
      AdaBrowse+ &	0.05 &	241	& 15.12 &	19.4	& 20.5	& 19.4 &	20.5 & 31.1 & 30.6 & 0.6 \\
      AdaBrowse+ & 0.10  & \textbf{171} (\color{blue}{$\downarrow$2.12$\times$})    & \textbf{17.58} (\color{blue}{$\uparrow$1.44$\times$}) & 19.6 &  20.7 & 19.5 & 20.6  & 31.2 & 30.7 & 0.7     \\
      \hline
    \end{tabular}
    \label{tab1}
  \end{table*}
  
\textbf{Implementation details.}
\textbf{Stage \uppercase\expandafter{\romannumeral1}}: The training procedure follows identical settings with recent CSLR methods~\cite{Min_2021_ICCV,cheng2020fully,pu2020boosting,niu2020stochastic} for fair comparison. ResNet18~\cite{he2016deep} with ImageNet~\cite{deng2009imagenet} pretrained weights is adopted for $f_L$ and $f_G$. The recognition model is consisted of 1D CNNs, a two-layer BiLSTM of hidden size 1024 and a fully connected layer, where sequential layers of \{K5, P2, K5, P2\} are adopted as 1D CNNs and $K\sigma $ and $P\sigma$ denotes a 1D convolutional layer and a max pooling layer with kernel size $\sigma$, respectively. Totally, we train models for 40 epochs with batch size 2 on a 3090 GPU. Adam optimizer is used with weight decay $10^{-4}$ and an initial learning rate of $10^{-4}$, divided by five after 20 and 30 epochs. Input frames are first resized to 256$\times$256 and randomly cropped to 224$\times$224 during training, augmented with 50\% random horizontal flip, and $\pm$20\% random temporal scaling. During testing, a center 224$\times$224 crop is simply adopted. Following VAC~\cite{Min_2021_ICCV}, two losses, i.e. VA Loss and VE Loss are added for additional visual supervision. The weight of VA Loss and VE Loss are set as 25.0 and 1.0 respectively.  We offer three choices ($M$=3) for subsequences, i.e. $\frac{1}{4}$, $\frac{1}{2}$ and 1, and three resolution candidates ($R$=3) , i.e. $224\times 224$, $160\times 160$ and $96\times 96$. Thus the number of all candidates is 7. 

\textbf{Stage \uppercase\expandafter{\romannumeral2}}: Generally, the training procedure follows similar settings with Stage \uppercase\expandafter{\romannumeral1}, excepting only training for 30 epochs with learning rate of $10^{-4}$, divided by five after 10 and 20 epochs, respectively. $\mathcal{L}_{{\rm Effi}}$ is employed to coordinate the internal components whose weight $\alpha$ is set within [0.04, 0.15] to produce a set of models to balance between accuracy and computations.

\textbf{Evaluation Metric.} 
We use word error rate (WER) to evaluate the performance of recognition, which is defined as the minimal summation of the \textbf{sub}stitution, \textbf{ins}ertion and \textbf{del}etion operations to convert the predicted sentence to the reference sentence as:
  \begin{equation}
    \label{e13}
    {\rm WER = \frac{\#sub+\#ins+\#del}{\#reference}.}
  \end{equation}
Note that the \textbf{lower} WER, the \textbf{better} accuracy.

\subsection{Analytical Results}
For clarity, we denote $\lambda R_{\eta}$ as a subsequence with $\lambda$ input length over resolution of $\eta \times \eta$. For example, $\frac{1}{4}R_{224}$ denotes a $\frac{1}{4}$ subsequence over resolution of 224$\times$224.  We use $R_{224}$ as our baseline. 

\textbf{Effectiveness.} We show the effectiveness of AdaBrowse and AdaBrowse+ on the PHOENIX14, PHOENIX14-T, CSL-Daily and CSL datasets in tab.~\ref{tab1}. Notably, our AdaBrowse generalizes well upon different datasets that are shot over various environments and expressed with different sign languages. Our baseline consumes 364G FLOPs of computations and achieves a throughput of 12.22 videos/s. By incorporating temporal feature redundancy into consideration, our proposed AdaBrowse notably decreases the required FLOPs to 254G and raises the throughput to 15.12 videos/s, retaining comparable accuracy. By further incorporating the spatial feature redundancy into the paradigm, our proposed AdaBrowse+ further increases the efficiency by reducing the FLOPs to 171G ($\downarrow$2.12$\times$) and promoting the throughput to 17.58 videos/s (1.44$\times$). $\mathcal{L}_{{\rm Effi}}$ controls the computation-WER trade-off. As the weight for $\mathcal{L}_{{\rm Effi}}$ increases, the WER is observed to degrade slightly with significantly reduced FLOPs and promoted throughput. 

\begin{table}[t]
  \setlength\tabcolsep{3pt}
  \centering
  \caption{Comparison of AdaBrowse with lightweight backbones on the PHOENIX14 dataset. }
  \begin{tabular}{cccc}
    \hline
    \multirow{2}{*}{Backbones} &\multirow{2}{*}{\makecell[c]{Throughput \\ (videos/s)}} & \multicolumn{2}{c}{PHOENIX14} \\
    &    & Dev(\%)    & Test(\%)  \\
    \hline
    Baseline(ResNet18~\cite{he2016deep}) & 12.22 & 19.7  & 20.9 \\
    RegNetX-800mf~\cite{radosavovic2020designing} & 12.55 &  19.9  & 21.1   \\
    SqueezeNet~\cite{iandola2016squeezenet}    & 13.13   & 21.6  & 22.3    \\
    ShuffleNet V2~\cite{ma2018shufflenet}  &  12.35  & 20.7  & 22.1    \\
    MobileNet V2~\cite{howard2018inverted}  &  12.35  & 21.3   & 22.5    \\
    EfficientNet B0~\cite{tan2019efficientnet}  &  12.35  & 21.4   & 22.5    \\
    \hline
    AdaBrowse  & 14.26  & 19.6 & 20.8 \\
    AdaBrowse+ & \textbf{17.58} & \textbf{19.6} &  \textbf{20.7}  \\
    \hline
  \end{tabular}
  \label{tab2}
\end{table}

\textbf{Comparison with lightweight backbones.}  Various lightweight CNNs have been proposed recently for efficiency. We compare AdaBrowse with some widely-used lightweight backbones by replacing $f_L$ with them in tab.~\ref{tab2}. It's observed that although these backbones could slightly promote the throughput, they mostly lead to a drastic accuracy drop. Our AdaBrowse and AdaBrowse+ not only achieve much better throughput but also outperform them on accuracy. 

\begin{table}[t]
  \setlength\tabcolsep{3pt}
  \centering
  \caption{Comparison of AdaBrowse with other adaptive methods on the PHOENIX14 dataset. }
  \begin{tabular}{ccccc}
  \hline
  \multirow{2}{*}{Backbones} & \multirow{2}{*}{GFLOPs} &\multirow{2}{*}{\makecell[c]{Throughput \\ (videos/s)}} & \multicolumn{2}{c}{PHOENIX14} \\
    &  &   & Dev(\%)    & Test(\%)  \\
  \hline
  Baseline	& 364	& 12.22	& 19.7	& 20.9 \\
  AdaFocus~\cite{wang2021adaptive}	& 232	& 13.62	& 21.0 &	22.1  \\
  Ar-net~\cite{meng2020ar}	& 254	& 13.87	& 20.2 	& 21.2  \\
  AdaFrame~\cite{wu2019adaframe}	& 217	& 14.12	& 23.6 	& 24.3  \\
  \hline
  AdaBrowse	& 254	& 14.26	& 19.6	& 20.8\\
  AdaBrowse+	& \textbf{171}	& \textbf{17.58} & \textbf{19.6} &  \textbf{20.7}  \\
\hline
\end{tabular}
\label{tab3}

\end{table}

  \begin{figure*}[t]
    \centering
    \begin{minipage}{0.33\textwidth}
    \centering
    \includegraphics[width=\textwidth]{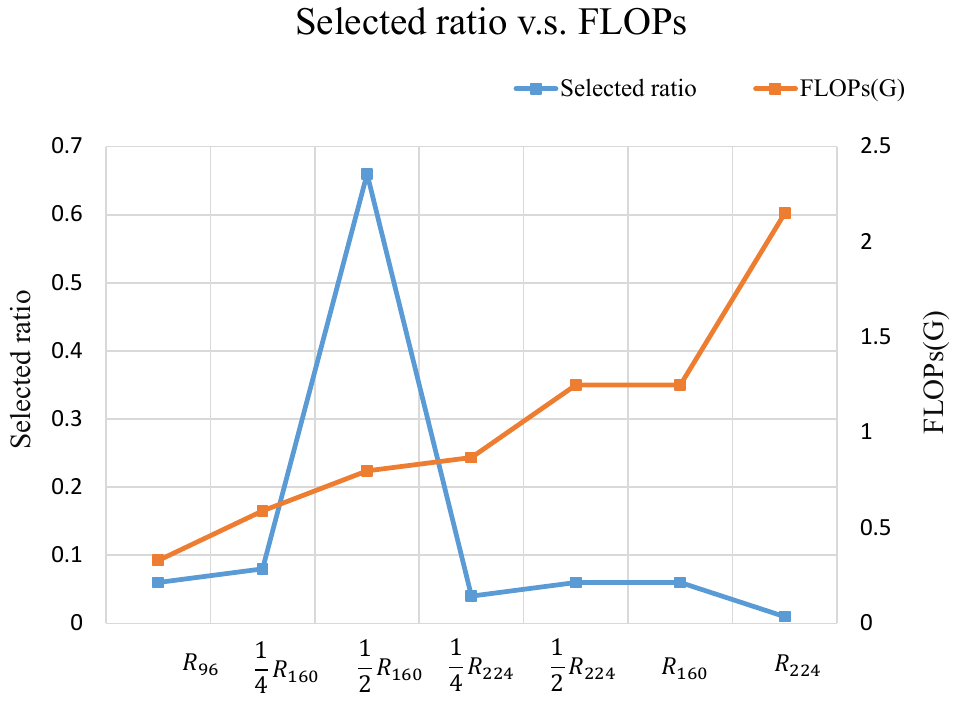}
    \caption{Selected ratio v.s. FLOPs for seven candidates of AdaBrowse+.}
    \label{fig5}
    \end{minipage} \hspace{0.03em}
    \begin{minipage}{0.32\textwidth} 
      \centering
      \includegraphics[width=\textwidth]{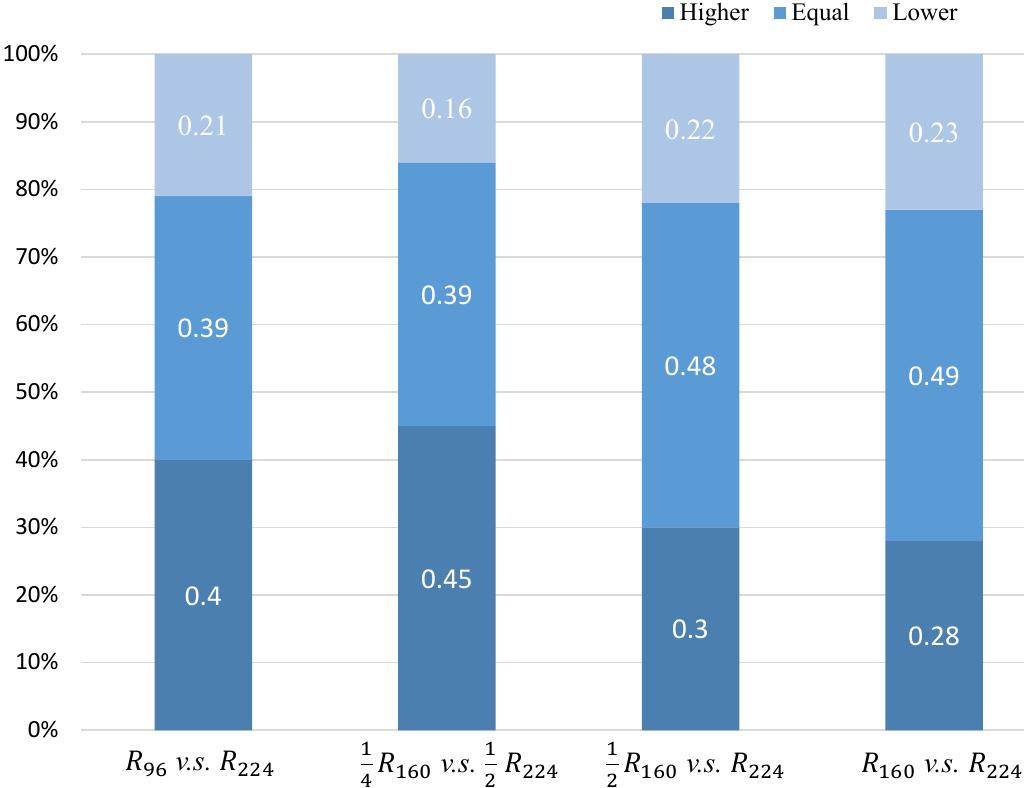}
      \caption{WER comparison over all input videos for some candidates against $R_{224}$.}
      \label{fig6}
      \end{minipage} \hspace{0.05em}
    \begin{minipage}{0.32\textwidth}
      \centering
      \includegraphics[width=\linewidth]{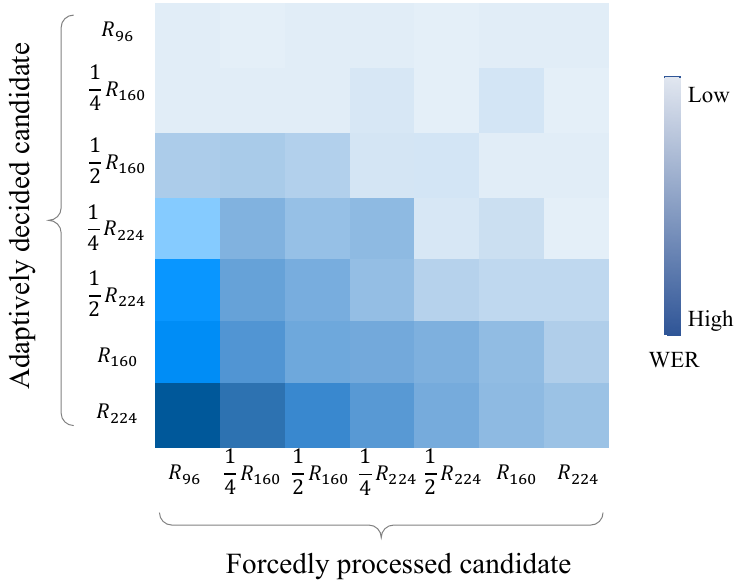}
      \caption{Illustration of WER with respect to processing order of candidates. }
      \label{fig7}
      \end{minipage}%
  \end{figure*}
 
\label{comp_ada}
\textbf{Comparison with other adaptive methods.} As no adaptive methods for CSLR have been explored before, we manually reproduce some approaches for CSLR to show the effectiveness of AdaBrowse. AdaFocus~\cite{wang2021adaptive} dynamically attends to the most informative small region in each frame to save unnecessary computations. Ar-net~\cite{meng2020ar} adaptively decides the frame resolution for each frame to avoid evenly allocating computing resources among frames. AdaFrame~\cite{wu2019adaframe} tries to only sample a portion of input frames for recognition and discard the rest. Notably, these methods mostly solely consider the temporal redundancy or the spatial redundancy. As shown in tab.~\ref{tab3}, these methods always lead to degraded accuracy. Our AdaBrowse series achieve much better accuracy than them with much higher throughput and fewer computations.

\textbf{Behavior analysis.} We are interested in how AdaBrowse internally makes decisions to balance between accuracy and computations. We plot the visualizations of the selected ratio v.s. FLOPs for each candidate in Adabrowse+ in fig.~\ref{fig5}. Candidates in the horizontal axis are arranged in reverse order (small$\mapsto $big) by their FLOPs over each video. It's observed that AdaBrowse+ tends to mainly (66\%) select $\frac{1}{2}R_{160}$ and almost evenly choose the rest. By incorporating spatial-temporal redundancy into consideration, AdaBrowse+ could dynamically choose to use a lower resolution (160$\times$160) than normal design with partial inputs ($\frac{1}{2}$), satisfying the need for accurate recognition. 
  
We further explore why AdaBrowse+ tends to make above decisions. We compare the WER between some candidates ($R_{96}$, $\frac{1}{4}R_{160}$ and $\frac{1}{2}R_{160}$) and the most expensive $R_{224}$ (which should give most accurate predictions) in fig.~\ref{fig6}. It's noticed that in \textgreater 60\% input samples, all these candidates can give equal or better predictions compared to $R_{224}$ with much fewer computations. $\frac{1}{2}R_{160}$ achieves the most superior trade-off. It gains competing accuracy compared to $R_{160}$ but consumes much fewer computations. This may result in its wide usage in decisions of AdaBrowse+. An extra detail is the chance of gaining equal or better predictions by $\frac{1}{2}R_{160}$ (70\%) compared to $R_{224}$ is quite close to its chance (66\%) selected by AdaBrowse+, demonstrating the intelligence over making decisions.

  \begin{figure}[t]
    \centering
    \includegraphics[width=\linewidth]{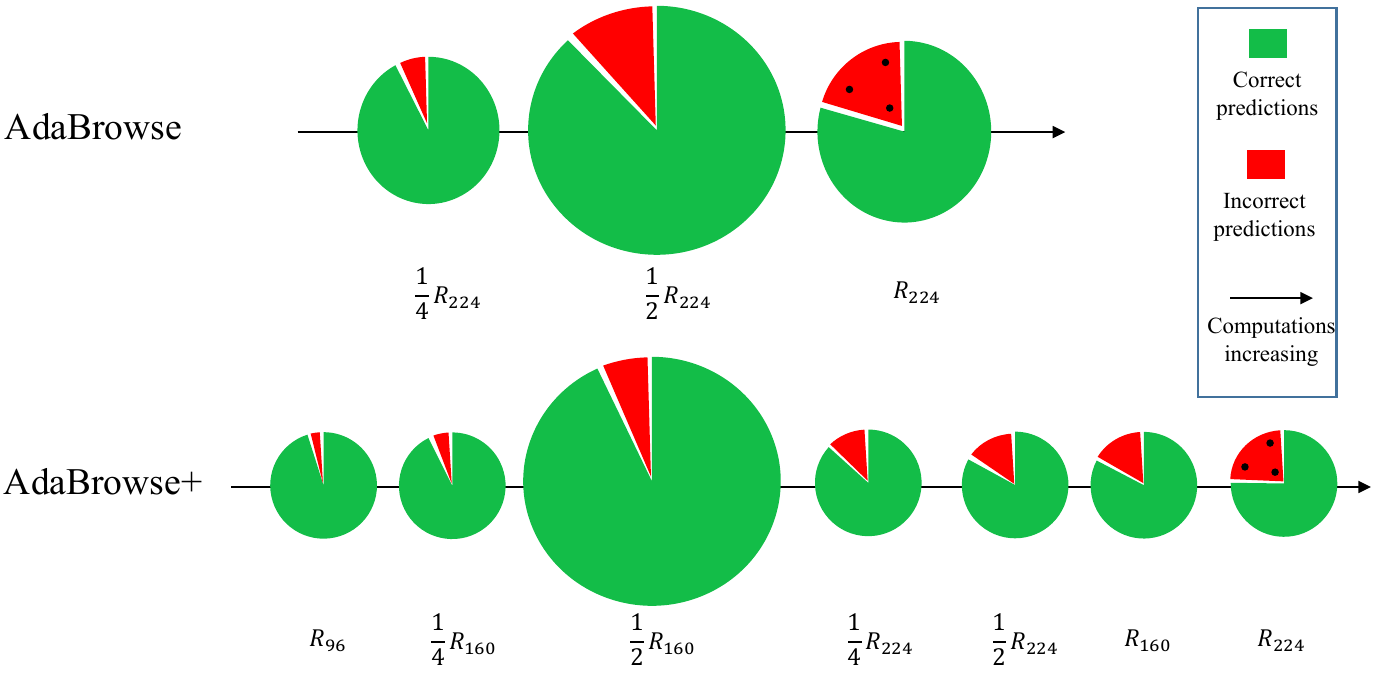} 
    \caption{Watermelon visualization for AdaBrowse and AdaBrowse+ over candidates. The area of a circle represents the percentage of processed videos for each candidate. Easier videos are processed by lightweight candidates with higher accuracy while only hard videos reach heavy candidates leading to increased misclassifications.}
    \label{fig8}
  \end{figure}
  
We further analyze the effectiveness of AdaBrowse in fig.~\ref{fig8} by illustrating the prediction accuracy v.s. the number of processed videos for each candidate in AdaBrowse and AdaBrowse+. A possibly counterintuitive observation is that as videos proceed to heavier candidates, the predictions become more inaccurate. However, the reason behind this is that easy videos have exited from earlier lightweight candidates, while only hard videos are left for heavy ones, thus leaving poorer accuracy. This phenomenon can be further verified in fig.~\ref{fig7} where the vertical axis denotes the candidate adaptively decided by AdaBrowse+ and the horizontal axis denotes the forcedly processed candidate for a video. Interestingly, we notice that if a video is processed by heavier candidates with more computations, no significant accuracy is gained. In contrast, if it's dealt with more lightweight candidates, a considerable accuracy drop is witnessed, which reflects the intelligence of AdaBrowse to allocate computing resources.
  
\begin{table*}[t]
  \centering
  \caption{Comparison with other methods over both WER and FLOPs on the PHOENIX14 and PHOENIX14-T datasets. $\star$ means additional factors such as face and hands are used. AdaBrowse and AdaBrowse+ achieve the best WER-computation trade-off.}
  \begin{tabular}{ccccccccc}
  \hline
  \multirow{3}{*}{Methods} &\multirow{3}{*}{Backbone}   &
  \multirow{3}{*}{\makecell{GFLOPs \\per video}} & \multicolumn{4}{c}{PHOENIX14} & \multicolumn{2}{c}{PHOENIX14-T} \\
  & & &\multicolumn{2}{c}{Dev(\%)} & \multicolumn{2}{c}{Test(\%)} &  \multirow{2}{*}{Dev(\%)} & \multirow{2}{*}{Test(\%)}\\
  & & & del/ins & WER & del/ins& WER & & \\
  \hline
  FCN~\cite{cheng2020fully}& Custom & 286 & - & 23.7 & -& 23.9 & 23.3& 25.1\\
  CMA~\cite{pu2020boosting} & GoogLeNet & 316 & 7.3/2.7 &21.3 & 7.3/2.4 & 21.9  & -&-\\
  VAC~\cite{Min_2021_ICCV}& ResNet18 & 367 & 7.9/2.5 & 21.2 &8.4/2.6 & 22.3 &- &-\\
  SMKD~\cite{hao2021self}& ResNet18 & 367 &6.8/2.5 &20.8 &6.3/2.3 & 21.0 & 20.8 & 22.4\\
  TLP~\cite{hu2022temporal} & ResNet18 & 367  & 6.3/2.8 & 19.7 & 6.1/2.9 & 20.8 & 19.4  & 21.2 \\
  SEN~\cite{hu2023self} & ResNet18 & 367 & 5.8/2.6 &  19.5 &  7.3/4.0 &  21.0 &  19.3 &  20.7 \\
  \hline
  SLT$^*$~\cite{camgoz2018neural}& GoogLeNet & 316 & - & - & - & - & 24.5 & 24.6\\
  CNN+HMM+LSTM$^*$~\cite{koller2019weakly}& GoogLeNet & 316 & - &26.0 & - & 26.0 & 22.1 & 24.1 \\
  STMC$^*$~\cite{zhou2020spatial}& VGG11 & 1554 & 7.7/3.4 &21.1 & 7.4/2.6 & 20.7 & 19.6 & 21.0\\
  C$^2$SLR$^*$~\cite{zuo2022c2slr} & ResNet18 & 367  & - & 20.5 &- & 20.4 & 20.2 & 20.4  \\
  \hline
  AdaBrowse  & ResNet18 &  254  &6.0/2.6   & 19.6 & 6.0/2.8 & 20.8 & 19.4 & 20.6        \\
  AdaBrowse+  & ResNet18 & \textbf{171} &6.0/2.5  & 19.6 &5.9/2.6  &  20.7 & 19.5 & 20.6        \\
  \hline
  \end{tabular}
  \label{tab5}
\end{table*}
  
\begin{table}[!t]   
	\centering
  \caption{Comparison with state-of-the-art methods on the CSL-Daily dataset~\cite{zhou2021improving}.} 
	\begin{tabular}{cccc}
	\hline
	Methods & \makecell{GFLOPs \\per video} &  Dev(\%) & Test(\%)\\
	\hline
	LS-HAN~\cite{huang2018video} & -  & 39.0  & 39.4\\
	TIN-Iterative~\cite{zhou2021improving} &  632 & 32.8  & 32.4\\ 
	Joint-SLRT~\cite{camgoz2020sign} & 316  & 33.1  & 32.0 \\
	FCN~\cite{cheng2020fully} & 286 & 33.2  & 32.5 \\
	BN-TIN~\cite{zhou2021improving} & 316 & 33.6  & 33.1 \\
	\hline
	AdaBrowse & 254 & 31.3 & 30.6 \\
	AdaBrowse+& \textbf{171} & 31.2 & 30.7 \\
	\hline
	\end{tabular}  
	\label{tab6}
	\end{table}
	
\begin{table}[!t]   
	\centering
  \caption{Comparison with state-of-the-art methods on the CSL dataset~\cite{huang2018video}.} 
	\begin{tabular}{ccc}
	  \hline
	  Methods& \makecell{GFLOPs \\per video}&  WER(\%)\\
	  \hline
	  SubUNet~\cite{cihan2017subunets} & -   & 11.0\\
	  SF-Net~\cite{yang2019sf} & 364 & 3.8 \\
	  FCN~\cite{cheng2020fully} & 286  & 3.0 \\
	  STMC~\cite{zhou2020spatial} & 1554 & 2.1 \\
	  VAC~\cite{Min_2021_ICCV} & 364 & 1.6 \\
	  \hline
	  AdaBrowse & 254 & 0.8 \\
	  AdaBrowse+ & \textbf{171} & 0.7 \\
	  \hline
	  \end{tabular}  
	  \label{tab7}
	\end{table}

\textbf{Comparison with the state-of-the-art.} Tab.~\ref{tab7}, tab.~\ref{tab8} and tab.~\ref{tab9} compare our model with state-of-the-art methods~\cite{Min_2021_ICCV,pu2019iterative,niu2020stochastic,zhou2020spatial,cui2019deep,cheng2020fully,pu2020boosting,koller2017re,niu2020stochastic} over both WER and FLOPs. AdaBrowse and AdaBrowse+ not only achieve state-of-the-art accuracy, but also significantly lower the required GFLOPs, which achieves the best WER-computation trade-off. 
Notably, AdaBrowse series reduce required computations to only half compared to existing methods. 
Hopefully, AdaBrowse makes a step further towards real-life deployment of sign language understanding.

\subsection{Ablation Study}
\textbf{Effectiveness of adaptive policy} is verified in tab.~\ref{tab8} against random sampling and sampling from a Gaussian distribution. It's observed our adaptive policy outperforms these counterparts by a large margin over accuracy, demonstrating its effectiveness. 
  
\textbf{Effects of reusing $x_G$ for recognition} is ablated in tab.~\ref{tab9}, where only using $x_L$ for recognition degrades the accuracy by about 0.4\%-0.5\%. Two-branch fusion could take advantage of features from different hierarchies.

\textbf{Effects of $\mathcal{L}_{{\rm Align}}$} is verified in tab.~\ref{tab10}. $\beta=0$ means $\mathcal{L}_{{\rm Align}}$ is disabled. It's observed that as $\beta$ increases, AdaBrowse achieves better accuracy as well. Thus $\mathcal{L}_{{\rm Align}}$ could help $f_G$ to obtain more representative features by mimicking $f_L$. $\beta$ is set as 25.0 by default.

\begin{table}[t]   
      \centering
      \caption{Effects of adaptive policy of AdaBrowse+. }
      \begin{tabular}{ccccc}
        \hline
        \multirow{2}{*}{Methods}  & \multicolumn{2}{c}{PHOENIX14} & \multicolumn{2}{c}{PHOENIX14-T}\\
        & Dev(\%)       & Test(\%) & Dev(\%)       & Test(\%)      \\
        \hline
        Adaptive    & \textbf{19.6} & \textbf{20.7}  & \textbf{19.5} & \textbf{20.6}  \\
        Random      & 20.7          & 22.9    & 20.9         & 23.1       \\
        Gaussian    & 20.6        & 23.1    & 21.1         & 23.0      \\
        \hline
      \end{tabular}
      \label{tab8}
    \end{table}

  \begin{table}[t]   
      \centering    
      \caption{Effects of reusing $x_G$ for recognition.}  
      \begin{tabular}{ccccc}
      \hline
      \multirow{2}{*}{Reusing $x_G$}  & \multicolumn{2}{c}{PHOENIX14} & \multicolumn{2}{c}{PHOENIX14-T}\\
        & Dev(\%)       & Test(\%) & Dev(\%)       & Test(\%)      \\
      \hline
      \ding{53} & 20.1 & 21.2 & 20.0 & 21.2\\
      $\checkmark$ & \textbf{19.6} & \textbf{20.7}  & \textbf{19.5} & \textbf{20.6} \\
      \hline
      \end{tabular}
      \label{tab9}
      \vspace{-7px}
  \end{table}
  
  \begin{table}[t]   
    \centering      
    \caption{Effects of $\mathcal{L}_{{\rm Align}}$ for recognition.}
    \begin{tabular}{cccccccc}
    \hline
    $\beta$  & 0.0 & 10.0 & 20.0 & 25.0 & 30.0 & 50.0\\
    Dev(\%) & 20.1 & 19.9 &  19.7 & \textbf{19.6} & 19.8 & 20.0  \\
    Test(\%) & 21.2 & 21.1 & 20.9 & \textbf{20.7} & 20.8 & 21.0 \\
    \hline
    \end{tabular}
    \label{tab10}
    \vspace{-7px}
\end{table}

\subsection{Online Setting}
Consider the online scenario in the real-world scenarios, where a stream of frames comes in sequentially and the model may need to
output the sentence prediction at any time. We test our AdaBrowse in this case to show its effectiveness in the real-world applications. Specifically, we follow FCN~\cite{cheng2020fully} to build a fully convolutional network as a baseline, which only watches information from the current and past timesteps to make sentence predictions, and deploy our AdaBrowse upon it. We use video frames from the PHOENIX14 dataset to mimic the shot videos in the real life. Tab.~\ref{tab_online} shows the results. It's observed that compared to the baseline, our AdaBrowse could effectively promote the number of processed videos to 1.41$\times$ with 2.06 fewer FLOPs. Meanwhile, our AdaBrowse could achieve slightly better accuracy than the baseline. These results verify the effectiveness of our AdaBrowse in applications of real-world scenario, which may bridge the communication gap of hearing-impaired people and the hearing people.

\begin{table}[t]
  \setlength\tabcolsep{3pt}
  \caption{Effectiveness of AdaBrowse in the online setting. }
  \label{tab_online}
  \centering
  \begin{tabular}{cccc}
  \hline
  Methods  & Throughput & GFLOPs & WER  \\
  \hline
  Baseline	& 11.24	& 361	& 21.0	 \\
  AdaBrowse & \textbf{15.84 ($\uparrow$1.41$\times$)} &  \textbf{175 (2.06$\downarrow$)}      &\textbf{20.8} \\
\hline
\end{tabular}
\vspace{-7px}
\end{table}

\section{Conclusion}
We propose AdaBrowse towards efficient CSLR by dynamically selecting a most informative subsequence over multiple resolutions. Considerable efficiency is achieved by considering the temporal redundancy and is further promoted by incorporating spatial redundancy, with 1.44$\times$ throughput over 2.12$\times$ less FLOPs.
\begin{acks}
This work is supported by the National Natural Science Foundation of China (Project No. 62072334).
\end{acks}

\bibliographystyle{ACM-Reference-Format}
\bibliography{ref}

\appendix

\end{document}